\documentclass[sigconf]{acmart}

\usepackage{graphicx}
\usepackage{algorithm}
\usepackage{algorithmic}
\usepackage{subfigure}
\usepackage{array}
\usepackage{setspace}
\usepackage{float}
\usepackage{url}
\usepackage{balance}
\usepackage{amsfonts}
\usepackage{amsmath}
\usepackage{rotating}
\usepackage{multirow}
\usepackage{color}
\setcopyright{acmcopyright}

\copyrightyear{2017}
\acmYear{2017}
\setcopyright{acmcopyright}
\acmConference{CIKM'17}{November 6--10, 2017}{Singapore, Singapore}
\acmPrice{15.00}
\acmDOI{10.1145/3132847.3133067}
\acmISBN{978-1-4503-4918-5/17/11}

\begin{document}
\title{RATE: Overcoming Noise and Sparsity of Textual Features in Real-Time Location Estimation}

\author{Yu Zhang$^1$, Wei Wei$^2$, Binxuan Huang$^2$, Kathleen M. Carley$^2$, Yan Zhang$^1$}
\affiliation{%
  \institution{$^1$Key Laboratory of Machine Perception (MOE), Peking University, Beijing, China}
  \streetaddress{$^2$School of Computer Science, Carnegie Mellon University, Pittsburgh, USA}
  \city{yuz9@illinois.edu, \{weiwei, binxuanh, kathleen.carley\}@cs.cmu.edu, zhy@cis.pku.edu.cn}
}

\begin{abstract}
Real-time location inference of social media users is the fundamental of some spatial applications such as localized search and event detection. While tweet text is the most commonly used feature in location estimation, most of the prior works suffer from either the noise or the sparsity of textual features. In this paper, we aim to tackle these two problems. We use topic modeling as a building block to characterize the geographic topic variation and lexical variation so that ``one-hot" encoding vectors will no longer be directly used. We also incorporate other features which can be extracted through the Twitter streaming API to overcome the noise problem. Experimental results show that our \texttt{RATE} algorithm outperforms several benchmark methods, both in the precision of region classification and the mean distance error of latitude and longitude regression.
\end{abstract}



\maketitle

\begin{spacing}{0.99}

\section{Introduction}
Micro-blogging services such as Twitter, Tumblr and Weibo are regarded as indispensable platforms for information sharing and social networking. In recent years, estimating the location information of social media users has become a popular topic with some important applications. For example, the ability to select a group of users in the specific spatial range can enable analysis on real-time disaster information, localized friendship recommendation or investigations on the geographic variation in habits.

For Twitter users, while tweet text is the most commonly used feature in location inference, most of the prior works suffer from the following two problems of textual features.

\noindent \underline{\emph{Noise}}. Due to the length limitation of a tweet, there are always lots of non-standard usages of the tweeting language including abbreviations, typos, and emoji.

\vspace{1mm}

\noindent \underline{\emph{Sparsity}}. In contrast with the entire corpus, there is a very small proportion of words appearing in each short tweet. Therefore, the ``one-hot" encoding vectors of each tweet will be sparse and hard to deal with.

\vspace{1mm}

In this paper, we propose \texttt{RATE} to overcome noise and sparsity of textual features in ReAl-Time location Estimation.

To tackle the noise problem, we incorporate other information available from the retrieved tweet. Figure 1 shows an example of a tweet with 8 metadata extracted through the Twitter streaming API \cite{zubi16}. Note that the exact latitude and longitude can be extracted only if users open their GPS service. But few users choose to do so for the concern of privacy. The absence of GPS signals forces us to rely on other information.\footnote{In this paper, we adopt the same scenario as \cite{zubi16}. We want to solve the location estimation problem for real-time Twitter streams. In this scenario, it becomes infeasible to retrieve follower-followee relationships or to make plenty of queries to an access-limited database. Therefore, we cannot rely on social connections or some third-party information although it is easy to put them in the model from the technical perspective.}

\begin{figure*}
\begin{minipage}[t]{0.64\textwidth}
\centering
\includegraphics[height=1.76in]{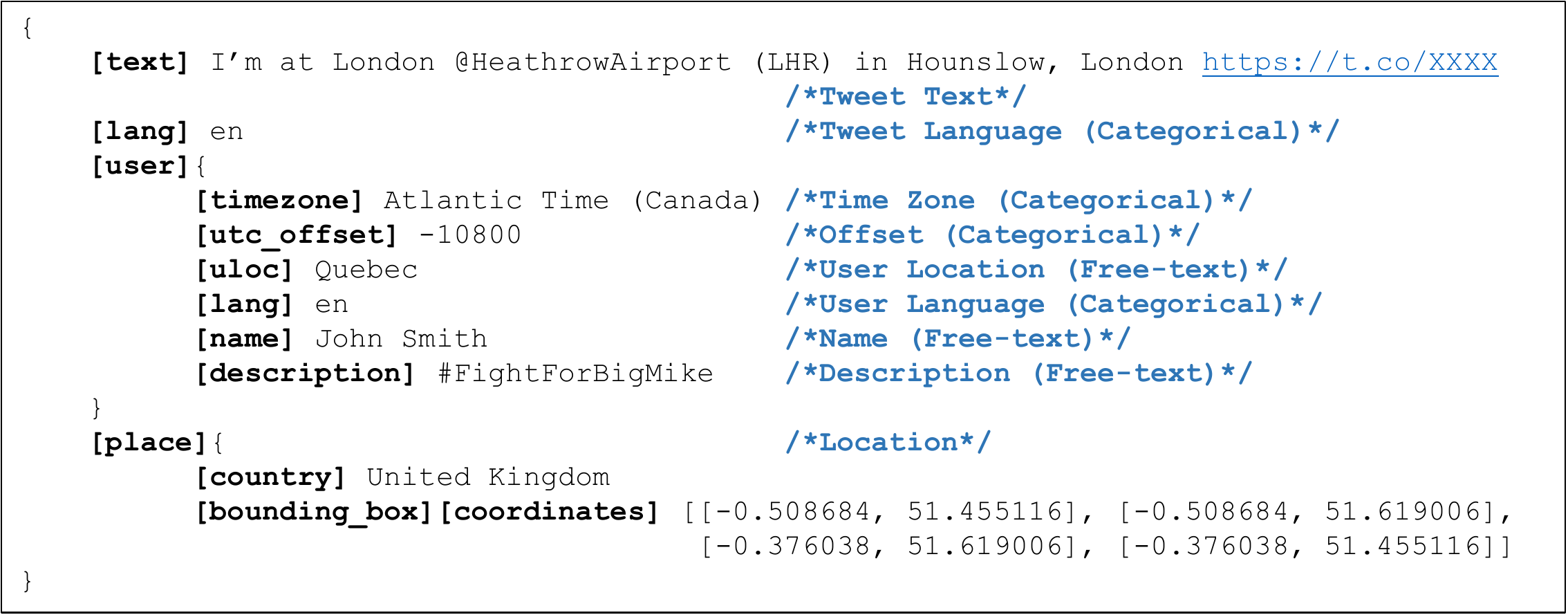}
\caption{Example of a geotagged tweet with 8 metadata \cite{zubi16}}
\label{fig:side:a}
\end{minipage}
\begin{minipage}[t]{0.35\textwidth}
\centering
\includegraphics[height=1.78in]{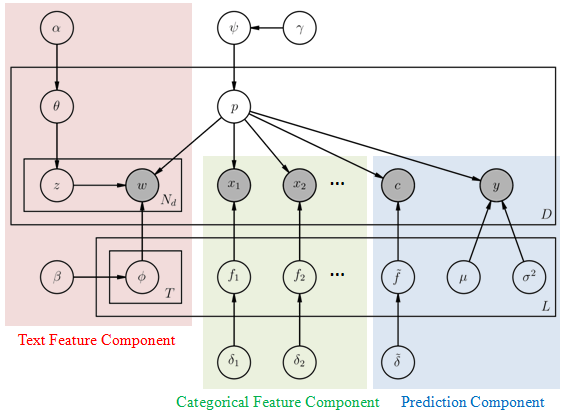}
\caption{Plate diagram of \texttt{RATE}}
\label{fig:side:b}
\end{minipage}
\end{figure*}

In Figure 1, user's residence, name and description are free-text fields completed during the registration. We can combine tweet text and some free-text fields into a single ``document". We name it as \underline{\emph{textual features}}. Besides, we also have several \underline{\emph{categorical features}} including time zone, UTC offset, tweet language and user language.

To tackle the sparsity problem, we use Latent Dirichlet Allocation \cite{blei02} as a building block to deal with textual features so that ``one-hot" encoding vectors will no longer be directly used as input.

Our generative model assumes that each location has different distributions over topics, words and categorical features. Therefore, we can infer the location through the observable features. Our model can be applied in either region classification or latitude and longitude regression, and experimental results show that \texttt{RATE} outperforms several benchmark methods in both tasks. As a byproduct, it can also be used to discover the real-time hot topics and find lexical variation of different regions.

\vspace{1mm}

\noindent \textbf{Related Work.} There is a long sequence of studies in location estimation of social media users. For more details, please refer to \cite{han14}.

Among various approaches used in location inference, finding location indication words is the most common one. For example, Cheng et al. \cite{cheng10} propose a local word filtering method. Eisenstein et al. \cite{eisen10} incorporate Correlated Topic Models (CTM) \cite{blei06c} to describe the relationships among different local topics. Chen et al. \cite{chen13} further add user's interests into their topic model.

Another widely used technique relies on social network relationships, so as to infer a user's location from that of its followers and followees \cite{sadilek12,jurgens13,dong14}. But in our context, a classifier needs to deal with the additional challenge of having to rely only on the information in a single tweet.

Our method is inspired by Zubiaga et al.'s work \cite{zubi16}. They use tweet metadata to train a Maximum Entropy classifier (i.e., logistic regression). However, they adopt ``one-hot" representation and ignore the sparsity of textual features.

\section{Model}

We use a Bayesian graphical model \texttt{RATE} to characterize the relationship between tweets, topics and regions. Using plate notation, Figure 2 illustrates the structure of our model.

There are $L$ latent areas. Each area $i$ has $T$ topics $\phi_{i,j}$ (represented by multinomial distributions over the words), $F$ categorical features $f_{u,i}$ (represented by multinomial distributions over the categories) and a region distribution $\tilde{f}_i$. As normal practice, we suppose that all multinomial distributions have a Dirichlet prior. We also assume that the latitude and longitude are extracted from a two-dimensional Gaussian distribution governed by the area's geographical center $\mu_i$ and variance $\sigma_i^2$. Given the area indicator $p$, we use the distributions of $p$ to generate words, categorical features, the region indicator and coordinates of the tweet.

\texttt{RATE} has three major components: the textual feature component, the categorical feature component and the prediction component, which have been marked in Figure 2.

The textual feature component has a similar structure with Latent Dirichlet Allocation (LDA) \cite{blei02}. The only difference is that in LDA, the word is selected by a per-token hidden variable $z$, while in \texttt{RATE}, the word is selected jointly by a topic index $z$ and a per-tweet area index $p$.

In the categorical feature component, features such as user language, tweet language and time zone are generated by the multinomial distribution $f_{u,i}$.

The prediction component includes the region indicator $c_i$ and the coordinates $y_i$ of the tweet. In the training set, they are the features which help us cluster the tweets and infer the latent parameters. And in the testing set, they are no longer observable and are the variables we want to predict.

We conclude our generative story as follows:

\begin{algorithm}
\vspace{-1mm}
\caption{Generative Process for \texttt{RATE}}
\label{alg:Framwork}
\begin{algorithmic}[1]
\STATE 1. Sample the distribution over areas $\psi \sim \text{Dir}(\gamma)$
\STATE 2. for each area $i = 1,2,...,L$
\STATE \ \ \ \ (1) for each topic $j = 1,2,...,T$
\STATE \ \ \ \ \ \ \ \ Sample the distribution over words $\phi_{i,j} \sim \text{Dir}(\beta)$
\STATE \ \ \ \ (2) for each categorical feature $u = 1,2,...,F$
\STATE \ \ \ \ \ \ \ \ Sample the distribution over categories $f_{u,i} \sim \text{Dir}(\delta_u)$
\STATE \ \ \ \ (3) Sample the distribution over regions $\tilde{f}_i\sim \text{Dir}(\tilde{\delta}_i)$
\STATE \ \ \ \ (4) Sample area center and variance $\mu_i \sim N(a, b^2I)$,
\STATE \ \ \ \ \ \ \ \ $\sigma^2_i\sim \Gamma(c,d)$
\STATE 3. for each document $k = 1,2,...,D$
\STATE \ \ \ \ (1) Sample area indicator $p_k \sim \text{Mul}(\psi)$
\STATE \ \ \ \ (2) Sample the distribution over topics $\theta_k \sim \text{Dir}(\alpha)$
\STATE \ \ \ \ (3) for each word position $l = 1,2,...,N_k$
\STATE \ \ \ \ \ \ \ \ Sample topic indicator $z_{k,l} \sim \text{Mul}(\theta_k)$
\STATE \ \ \ \ \ \ \ \ Sample word $w_{k,l} \sim \text{Mul}(\phi_{z_{k,l},p_k})$
\STATE \ \ \ \ (4) for each categorical feature $u = 1,2,...,F$
\STATE \ \ \ \ \ \ \ \ Sample category indicator $x_{u,k} \sim \text{Mul}(f_{u,p_k})$
\STATE \ \ \ \ (5) Sample region indicator $c_k \sim \text{Mul}(\tilde{f}_{p_k})$
\STATE \ \ \ \ (6) Sample coordinates $y_k \sim N(\mu_{p_k}, \sigma^2_{p_k}I)$
\end{algorithmic}
\end{algorithm}

As common sense, different regions may have different popular topics. Even for the same topic, there exists a geographic lexical variation \cite{eisen10}. Different regions may also have distinct distributions of user's features. For example, French will be the dominating element of the user language distribution in France, while it will not cover a considerable proportion in Germany.  Therefore, the observable text and user features of the tweet are strong spatial indicators.

\begin{table*}
\centering
\caption{Location prediction results.}
\vspace{-1mm}
\begin{tabular}{|c|ccc|cc|cc|}
\hline
Dataset     & \multicolumn{3}{c|}{\texttt{Europe}} &  \multicolumn{2}{c|}{\texttt{UK}} &  \multicolumn{2}{c|}{\texttt{France}} \\
\hline
Method	    &  Precision &	MDE(km)	 & Time(ms)	& Precision	& MDE(km)	             & Precision	& MDE(km) \\
\hline
\texttt{NB}	        &  0.8770 	 & 427.0 	 & 0.20 	& 0.3842 	& 215.0 	             & 0.5218 	    & 284.6   \\
\texttt{SVM}     	&  0.8796 	 & 426.3 	 & 3.56 	& 0.4188 	& 205.8 		         & 0.5395       & \textbf{275.8}   \\
\texttt{GeoTM}      &  0.7973 	 & 529.8 	 & 16.5 	& 0.4260 	& 195.6	                 & 0.5414       & 277.9   \\
\texttt{LR-Text} 	&  0.7229 	 & 726.9 	 & 2.48 	& 0.3983 	& 224.6 		         & 0.5316       & 310.9   \\
\texttt{LR-Full} 	&  0.8890 	 & 429.2 	 & 2.46 	& 0.4207 	& 214.6 	             & 0.5413 	    & 285.4   \\
\hline
\texttt{RATE}	&  \textbf{0.8922}  & \textbf{372.8}  & 17.5  & \textbf{0.4325}  & \textbf{194.8}  & \textbf{0.5586}  & 284.9   \\ 	
\hline
\end{tabular}
\end{table*}

\vspace{1mm}

\noindent \textbf{Inference.} We use a Gibbs-EM algorithm \cite{wallach06} to infer the model parameters. During the E step, we assume that $\mu$ and $\sigma^2$ are already known as the result of a previous M step. We then use Collapsed Gibbs Sampling to generate samples for $z$ and $p$ and use the average of these samples to approximate the expectation:
\begin{equation}
\begin{split}
\Pr(z_{kl} = z|\neg z_{kl}) \propto (n_{k,*}^{z,*-}+\alpha_z)\cdot\frac{n_{*,r}^{z,p_k-}+\beta_r}{\sum_{r=1}^V(n_{*,r}^{z,p_k-}+\beta_r)}, \notag
\end{split}
\end{equation}
and
\begin{equation}
\begin{split}
&\Pr(p_k = p|\neg p_k)\propto \frac{1}{\sigma^2_p}\exp(-\frac{||y_k-\mu_p||_2^2}{2\sigma^2_p}) \cdot \prod_{l=0}^{N_k-1}(n_{*,*}^{*,p-}+\gamma_p+l) \cdot \\
&\prod_{j,r:n_{k,r}^{j,*}>0}\frac{\prod_{l=0}^{n_{k,r}^{j,*}-1}(n_{*,r}^{j,p-}+\beta_r+l)}{ \prod_{l=0}^{n_{k,*}^{j,*}-1}(\sum_{r=1}^V(n_{*,r}^{j,p-}+\beta_r)+l)}\cdot \prod_{u=1}^{F+1}\frac{m_{*,u}^{p,x_{ku}-}+\delta_{u,x_{ku}}}{\sum_{v=1}^{C_u}(m_{*,u}^{p,v-}+\delta_{u,v})}. \notag
\end{split}
\end{equation}
Here $n_{k,r}^{j,i}$ denotes the number of times that a document $k$ has a word $r$ that falls into topic $j$ in area $i$, and $m_{k,u}^{i,v}$ is the number of times that a feature $u$ of document $k$ falls into category $v$ of area $i$. Note that $c_i$ and $y_i$ are observable in the training set. Therefore, we regard $c_i$ as the $(F+1)$-th categorical feature of tweet $i$.

In the M step, we estimate $\mu$ and $\sigma^2$ by maximizing the likelihood function, which is defined as the average over all samples drawn from the E step:
\begin{equation}
\begin{split}
Q(\mu, \sigma^2) = \frac{1}{S}\sum_{s=1}^S\log(\Pr(O,\Omega^{(s)}|\mu,\sigma^2))-\frac{1}{2}\lambda||\sigma||_2^2. \notag
\end{split}
\end{equation}

By solving the equations $\frac{\partial Q}{\partial \mu_p} = 0$ and $\frac{\partial Q}{\partial \sigma_p} = 0$, we acquire an MLE estimation for the center and variance of each region:
\begin{equation}
\begin{split}
\mu_p = \frac{\sum_{s=1}^S\sum_{k:p_k^{(s)}=p}y_k}{\sum_{s=1}^S\sum_{k:p_k^{(s)}=p}1}, \notag
\end{split}
\end{equation}
and $\sigma_p^2$ is the positive root of the following biquadratic equation
\begin{equation}
\begin{split}
\sum_{s=1}^S\sum_{k:p_k^{(s)}=p}(\lambda\sigma_p^4+\sigma_p^2-\frac{1}{3}||y_k-\mu_p||_2^2) = 0. \notag
\end{split}
\end{equation}

For a tweet in the test set, we can either make a point estimation on the latitude and longitude or make a region classification.

For latitude and longitude regression, we have
\begin{equation}
\hat{y}_k = \frac{\sum_{s=1}^S\mu_{p_k^{(s)}}/\sigma^2_{p_k^{(s)}}}{\sum_{s=1}^S1/\sigma^2_{p_k^{(s)}}}. \notag
\end{equation}

For the classification, we have
\begin{equation}
\hat{c}_k = \textrm{arg}\max_C \prod_{s=1}^S \tilde{f}_{p_k^{(s)},C}. \notag
\end{equation}

\section{Experiments}
\noindent \textbf{Dataset.} We extract a Twitter dataset within the geographical boundary of Europe from October 2015 to December 2015. The boundary is defined by the (latitude, longitude) point (-13.97, 33.81) in the lower-left corner and (41.40, 58.73) in the upper right corner. We remove the users who posted less than 10 tweets during these 3 months and get 376,356 users left. To avoid bias in the dataset, we randomly select one tweet for each user. We name this dataset as \texttt{Europe}.

In \texttt{Europe}, we select all the tweets from UK/France to form another 2 datasets. After the filtering, we have 77,852 and 36,451 tweets in \texttt{UK} and \texttt{France} respectively.

In all of the 3 datasets, we use 60\% of tweets for training, 20\% for tuning the parameters, and the remaining 20\% for final testing. For the tweet text, we remove URLs and the words that occur less than 10 times in the whole corpus. But we retain mentions (``$@$username"), hashtags (``$\#$topic") and stop words. After the preprocessing, we have a vocabulary of approximately 30K words.

For each tweet, we combine tweet text and the user's profile location into a single ``document". It is our textual feature. Besides, we use user language, time zone and tweet language as categorical features.

\vspace{1mm}

\noindent \textbf{Evaluation Metrics.} For the region classification task, we use \underline{\emph{precision}} to evaluate the performance, which is defined as the percent of tweets which are predicted in the same region where they are published. Note that in \texttt{Europe}, we directly conduct country-level classification. In \texttt{UK} and \texttt{France}, we divide each country into 4 regions according to the result of K-means.

For the coordinates regression task, we use \underline{\emph{mean distance}} \underline{\emph{error}} (MDE) as our metric. It is the average error distance (on the sphere) between predicted location and actual location.
%

\vspace{1mm}

\noindent \textbf{Effectiveness.} We select the following benchmark methods, which are also applicable in the real-time scenario, to compare with our approach.

\vspace{1mm}

(1) Naive Bayes (\texttt{NB}) is a basic classification method using only categorical features.

(2) \texttt{SVM} trains a linear Support Vector Machine with both categorical features and textual features.

(3) \texttt{LR-Text} \cite{zubi16} trains a Logistic Regression classifier using only textual features with ``one-hot" representation.

(4) \texttt{LR-Full} \cite{zubi16} is similar to \texttt{LR-Text}, but it incorporates both categorical features and textual features. According to the original paper, this combination performs the best.

(5) \texttt{RATE} is the method proposed in this paper.\footnote{The code as well as the dataset is available at \\ \texttt{https://github.com/yuzhimanhua/Location-Inference/}.}

(6) \texttt{GeoTM} is a simplification of \texttt{RATE}, using only textual features. It is also similar with the method in \cite{eisen10}. The only difference is that Eisenstein et al. use CTM, while we use LDA.

\vspace{1mm}

Table 1 shows the location prediction results of the methods mentioned above. As expected, \texttt{RATE} significantly outperforms all the benchmark methods, both in region classification and coordinates regression.

Methods only using textual features, such as \texttt{GeoTM} and \texttt{LR-Text}, perform not so well in country-level classification because categorical features do help a lot in dealing with the noise problem in coarse-grained tasks. However, textual features show their power in fine-grained tasks. We can see that \texttt{GeoTM} performs the second best in \texttt{UK} and \texttt{France}. where categorical features may have less contribution in location estimation. Therefore, if we want to balance the performances of our algorithm in both coarse-grained tasks and fine-grained ones, it will be effective to incorporate both textual features and categorical features into our model. Moreover, we should note that \texttt{RATE} adopts a better way than \texttt{SVM} and \texttt{LR-Full} in dealing with the sparsity of textual features.

\vspace{1mm}

\noindent \textbf{Efficiency.} Table 1 also shows the running time each algorithm spends on each tweet in \texttt{Europe}. Note that we only calculate testing time and do not take the training phase into account. We can observe that \texttt{RATE}, \texttt{SVM} and \texttt{LR-Full} are almost at the same order of magnitude in efficiency. Since we only adopt the original Collapsed Gibbs Sampling method in \texttt{RATE}, we believe that \texttt{RATE} can be even faster with the help of some acceleration strategies of sampling \cite{likdd2014}.

\vspace{1mm}

\noindent \textbf{Parameter Study.}
As common practice, we set $\alpha$ to be $50/(LT)$ and other Dirichlet priors to be 0.01.

Figure 3 shows the MDE of \texttt{RATE} in \texttt{Europe} with different numbers of regions ($L$) and topics ($T$). We can observe that for each fixed $L$, the model always performs the best in the case $T=1$. Therefore we no longer need to sample $\theta$ and $z$, and the structure of $\beta$, $\phi$, $w$, $p$, $\psi$ and $\gamma$ will be identical to the DMM model \cite{yin14}, which has proved to be effective in dealing with short text like tweets \cite{yin14}.

The ``best" $L$ is 30 in \texttt{Europe}, which approximately equals to the number of countries.

%
%

\vspace{1mm}

\noindent \textbf{Words and Topics.} As a byproduct in the training process of \texttt{RATE}, we show the top 8 words in the top 5 regions in \texttt{Europe} in Table 2.
The top words can be divided into four categories: temporal words (e.g., ``today" and ``october"), location names (e.g., ``paris" and ``spain"), local characteristic words (e.g., ``rain" and ``wind" in Britain and ``love" in France) and hashtags. These four kinds of words correspond to time, locations, topics and events respectively.

\section{Conclusion}
In this paper, we propose a Bayesian graphical model to overcome the noise and sparsity problems in real-time location estimation on Twitter. The key ideas of our model are that: (1) we use the combination of text information and user profile information to tackle the noise problem. (2) we use topic modeling characterizing the geographic lexical variation to tackle the sparsity problem. Quantitative analysis justifies our model on several Twitter datasets by showing that our approach outperforms several benchmark methods. Qualitative analysis shows that our model is also useful in extracting location-relevant topics.

\begin{figure}[t]
\centering
\includegraphics[scale=0.64]{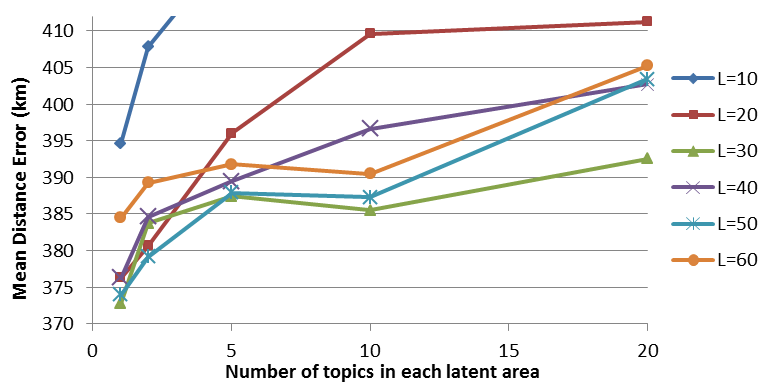}
\caption{Mean Distance Errors of \texttt{RATE} in \texttt{Europe} with different $L$ and $T$.}
\vspace{-0.5em}
\end{figure}

\begin{table}[t]
\centering
\caption{Top 8 words of the top 5 region in \texttt{Europe}. Stop words have been removed. The italic words in the brackets are the English translation.}
\vspace{-1mm}
\begin{tabular}{|c|c|}
\hline
 Location     & Word    \\
\hline
(36.48,31,36) & t\"{u}rkiye / istanbul / \"{u}niversitesi (\underline{\emph{university}}) \\
Turkey        & izmir / ankara / lisesi (\underline{\emph{high school}}) / antalya \\
              & fak\"{u}ltesi (\underline{\emph{faculty}}) \\
\hline
(56.19,-1.73) & london / night / wind / rain \\
UK            & tonight / october / year / life \\
\hline
(39.82,-3.49) & hoy (\underline{\emph{today}}) / madrid / ma\~{n}ana (\underline{\emph{morning}}) \\
Spain         & vida (\underline{\emph{lifetime}}) / spain / octubre (\underline{\emph{october}}) \\
              & \#gala4gh16 / jueves (\underline{\emph{thursday}}) \\
\hline
(47.69,2.76)  & paris / france / demain (\underline{\emph{tomorrow}}) \\
France        & mdr (\underline{\emph{lol}}) / soir (\underline{\emph{evening}}) / vie (\underline{\emph{life}}) \\
              & journ\'{e}e (\underline{\emph{day}}) / aime (\underline{\emph{love}}) \\
\hline
(43.66,11.03) & milano / italy / \#xf9 / italia \\
Italy         & \#gf14 / oggi (\underline{\emph{today}}) / roma / milan \\
\hline
\end{tabular}
\end{table}

\vspace{1mm}

\noindent\textbf{Acknowledgements.} We would like to thank Yujie Qian and Matthew Benigni for valuable discussions, and anonymous reviewers for useful feedback. This work is supported by NSFC under Grant No.61532001 and No.61370054.



\bibliography{cikm17}
\end{spacing}

\end{document}